\documentclass{elsart}
\journal{Fuzzy Sets and Systems}
\usepackage{graphicx}
\usepackage{cite}
\usepackage{latexsym}
\usepackage{times}
\usepackage{amsmath,amssymb}
\usepackage{algorithm}
\usepackage{algorithmic}
\usepackage{color}
\newtheorem{definition}{Definition}

\begin{document}

\begin{frontmatter}

\title{\Large\bf Building an interpretable fuzzy rule base from data
  using Orthogonal Least Squares \\Application to a depollution problem}

\author[INRA]{S\'ebastien~Destercke}, 
\author[Cemagref]{Serge Guillaume}, 
\author[INRA]{Brigitte Charnomordic\corauthref{cor}},
\corauth[cor]{Corresponding author.}
\ead{bch@ensam.inra.fr}

\address[Cemagref]{Umr Itap, Cemagref, BP 5095, 34196 Montpellier Cedex, France}
\address[INRA]{Umr ASB, INRA, 2 place Viala, 34060 Montpellier Cedex, France}

\normalsize

\begin{abstract}
In many fields where human understanding plays a crucial role, such as
bioprocesses, the capacity of extracting knowledge from data is of
critical importance. Within this framework, fuzzy learning methods,
if properly used, can greatly help human experts. Amongst
these methods, the aim of orthogonal transformations, which have been
proven to be mathematically robust, is to build rules from a set of
training data and to select the most important ones by linear
regression or rank revealing techniques. The OLS algorithm is a good
representative of those methods. However, it was originally designed so
that it only cared about numerical performance. Thus, we propose some
modifications of the original method to take interpretability into
account. After recalling the original algorithm, this paper presents
the changes made to the original method, then discusses some results
obtained from benchmark problems. Finally, the algorithm is applied to
a real-world fault detection depollution problem.
\end{abstract}

\begin{keyword}
Learning, rule induction, fuzzy logic, interpretability, OLS, orthogonal transformations, depollution, fault detection
\end{keyword}
\end{frontmatter}

\normalsize


\section{Introduction}
Fuzzy learning methods, unlike ``black-box'' models such as neural networks, are likely to give interpretable results, provided that some constraints are respected. While this ability is somewhat meaningless in some applications such as stock market prediction, it becomes essential when human experts want to gain insight into a complex problem (e.g. industrial \cite{Drobics04} and biological \cite{Woolf00} processes, climate evolution \cite{Han02}).

These considerations explain why interpretability issues in Fuzzy Modeling have become an important research topic, as shown in recent literature \cite{Casillas03}.
Even so, the meaning given to interpretability in Fuzzy Modeling is not always the same. By interpretability, some authors mean mathematical interpretability, as in \cite{Bikdash99} where a structure is developed in Takagi-Sugeno systems,
that leads to the interpretation of every consequent polynomial as a Taylor series expansion about the rule center. Others mean linguistic interpretability, as in \cite{Guillaume01b}, \cite{Glorennec03}. The present paper is focused on the latter approach.
Commonly admitted requirements for interpretability are a small number of consistent membership functions and a reasonable number of rules in the fuzzy system.

Orthogonal transformation methods provide a set of tools for building rules from data and selecting a limited subset of rules. Those methods were originally designed for linear optimization, but subject to some conditions they can be used in fuzzy models.
For instance, a zero order Takagi Sugeno model can be written as a set of r fuzzy rules, the $qth$ rule being:

\begin{equation}
R_q : \textrm{ if } x_1 \textrm{ is } A_1^q \textrm{ and } x_2
\textrm{ is } A_2^q \textrm{ and } \ldots \textrm{ then } y = \theta^q
\label{eq:rule}
\end{equation}

where $A_1^q, A_2^q \ldots$ are the fuzzy sets associated to the $x_1,x_2,\ldots$ variables for that given rule, and $\theta^q$ is the corresponding crisp rule conclusion.

Let $(x,y)$ be $N$ input-output pairs of a data set, where $x \in \mathbb{R}^p$ and $y \in \mathbb{R}$. For the $ith$ pair, the above Takagi Sugeno model output is calculated as follows:
\begin{equation}
\widehat{y^i} = \frac{\sum\limits_{q=1}^r {\theta^q \left ( \bigwedge\limits_{j=1}^p \mu_{A_j^q}(x_j^i) \right )}}{\sum\limits_{q=1}^r \left (\bigwedge\limits_{j=1}^p \mu_{A_j^q}(x_j^i) \right ) }
\label{eq:fuzzsys}
\end{equation}
In equation \ref{eq:fuzzsys}, $\bigwedge$ is the conjunction operator used to combine elements in the rule premise, $\mu_{A_j^q}(x_j^i)$ represents, within the $qth$ rule, the membership function value for $x^i_j,j=1 \ldots p$.

Let us introduce the rule firing strength $w^q(x^i)=\bigwedge\limits_{j=1}^p \mu_{A_j^q}(x_j^i)$.
Thus equation \ref{eq:fuzzsys} can be rewritten as:
\begin{equation}
\widehat{y^i} = \frac{\sum\limits_{q=1}^r {\theta^q w^q(x^i)}}{\sum\limits_{q=1}^r {w^q(x^i)}}
\label{eq:fuzzymatchdeg}
\end{equation}

Once the fuzzy partitions have been set, and provided a given data set, the $w^q(x^i)$ can be computed for all $x^i$ in the data set. Then equation \ref{eq:fuzzymatchdeg} allows to reformulate the fuzzy model as a linear regression problem, written in matrix form as:
$y = P \theta + E$. In that matrix form, y is the sample output vector, P is the firing strength matrix, $\theta$ is the rule consequent vector and E is an error term. Orthogonal transformation methods can then be used to determine the $\theta^q$ to be kept, and to assign them optimal values in order to design a zero order Takagi Sugeno model from the data set.

 A thorough review of the use of orthogonal transformation methods (SVD, QR, OLS) to select fuzzy rules  can be found in \cite{Yen99}. They can be divided into two main families: the methods that select rules using the $P$ matrix decomposition only, and others that also use the output $y$ to do a best fit.
The first family of methods (rank revealing techniques) is particularly interesting when the input fuzzy partitions include redundant or quasi redundant fuzzy sets.
The orthogonal least squares (OLS) technique belongs to the second family and allows a rule selection based on the rule respective contribution to the output inertia or variance. With respect to this criterion, it gives a good summary of the system to be modeled, which explains why it has been widely used in Statistics, and also why it is particularly suited for rule induction, as shown for instance in \cite{Setnes03}.

The aim of the present paper is to establish, by using the OLS method as an example, that orthogonal transformation results can be made interpretable, without suffering too much loss of accuracy. This is achieved by building interpretable fuzzy partitions  and by reducing the number of rule conclusions.
This turns orthogonal transformations into useful tools for modeling regression problems and extracting knowledge from data. Thus they are worth a careful study as there are few available techniques for achieving this double objective, contrary to knowledge induction in classification problems.

In section \ref{section:OLS}, we recall how the original OLS works. Section \ref{section:criteria} introduces the learning criteria that will be used in our modified OLS algorithm. Section \ref{section:modifs} presents the modifications necessary to respect the interpretability constraints.  In the next section, the modified algorithm is applied to benchmark problems, compared to the original one and to reference results found in the literature. A real-world application is presented and analyzed in section \ref{sec:lbe}. Finally we give some conclusions and perspectives for future work.

\section{Original OLS algorithm}
\label{section:OLS}
The OLS (orthogonal least squares) algorithm \cite{Chen89,Chen91} can be used in Fuzzy Modeling to make a rule selection using the same technique as in linear regression.  Wang and Mendel \cite{Wang92b} introduced the use of Fuzzy Basis Functions to map the input variables into a new linear space. We will recall the main steps used in the original algorithm.

\bigskip
\textbf{Rule construction}

First N rules are built, one from each pair in the data set. Hohensohn and Mendel \cite{Hohensohn94} proposed the following Gaussian membership function for the $jth$ dimension of the $ith$ rule.

\begin{equation}
\displaystyle \mu_{A_j^i}(u)=e^{\left\lbrack -\frac{1}{2} \left( \frac{(u-x_j^i)}{\sigma_j} \right)^2 \right\rbrack}
\label{eq:gaussian}
\end{equation}

with  $\sigma_j=s . \lbrack \max \limits_{i=1,2,\ldots,N} ( x_j^i) - \min \limits_{i=1,2,\ldots,N} (x_j^i)\rbrack $, $s$ being a scale factor whose value depends on the problem.

\bigskip
\textbf{Rule selection}

Once the membership functions have been built, the Fuzzy Inference System (FIS) optimization is done in two steps. The first step is non-linear and consists in fuzzy basis function (FBF) construction; the second step, which is linear, is the orthogonal least square application to the FBF.

A FBF $p^i(x^i)$ is the relative contribution of the $ith$ rule, built from the $ith$ example, to the inferred output:

$$p^i(x^i) = \frac{w^i(x^i) }{\sum\limits_{q=1}^N w^q(x^i) }$$

Thus the fuzzy system output (see equation \ref{eq:fuzzymatchdeg}) can be written and viewed as a linear combination: $ \widehat{y^i} =  \sum\limits_q {p^q(x^i)\  \theta^q} $, where $\theta^q \in \mathbb{R}$ are the parameters to optimize (they correspond to the rule conclusions). The system is equivalent to the matrix form $y = P \theta + E$, $y$ being the observed output while $E$ is the error term, supposed to be uncorrelated with the $p^i(x)$ or $P$.

The element $p_{ji}$ of the matrix $P$ represents the $ith$ rule  firing strength for the $jth$ pair, i.e. the $jth$ component of the $p^i$ vector.

The OLS procedure transforms the $p^i$ regressors into a set of orthogonal ones using the Gram-Schmidt procedure. The $P$ matrix can be decomposed into an orthogonal one, $M$, and an upper triangle one, $A$. \\The system becomes \mbox{$y = MA\theta + E$}. Let $g=A \theta$, then the orthogonal least square solution of the system is $ \displaystyle \widehat{g_i} = \frac{m_i^Ty}{m_i^Tm_i}$, $1 \le i \le r$ where $m_i$ is the $ith$ column of the orthogonal matrix $M$.

Optimal $\widehat{\theta}$ can be computed using the triangular system $A \widehat{\theta} = \widehat{g}$.

Thanks to the orthogonal characteristic of $M$, there is no covariance, hence vector (i.e. rule) individual contributions are additive. This property is used to select the rules. At each step, the algorithm selects the vector $m_i$ that maximizes the explained variance of the observed output $y$. The selection criterion is the following one:

$$ [xVar]_i = \frac {g_i^2\  m_i^T m_i}  {y^Ty} $$

The selection stops when the cumulated explained variance is satisfactory. This occurs at step $r \leq N$ when

\begin{equation}
\displaystyle  1 - \sum\limits_{i=1}^{r} [xVar]_i < \epsilon
\label{eq:thresh}
\end{equation}

$\epsilon$ being a threshold value (e.g. 0.01).

\bigskip
\textbf{Conclusion optimization}

As the selected $m_i$ still contain some information related to unselected rules, Hohensohn and Mendel  \cite{Hohensohn94} propose to run the algorithm a second time. No selection is made during this second pass, the aim being only to optimize the rule conclusions.

The original algorithm, as described here, results in models with a
good numerical accuracy. However, as we'll see later on, it has many
drawbacks when the objective is not only numerical accuracy but also
knowledge extraction.

\section{Learning criteria}\label{section:criteria}
Two numerical criteria are presented: the coverage index, based upon
an activation threshold, and the performance index. We will use them
to assess the overall system quality. The performance index is an
error based index that will allow us to measure the numerical
accuracy of our results, while the coverage index, together with the
activation threshold, will give us information related to the system completeness
with respect to the learning data.
To some extent, the coverage index reflects the potential quality of the
extracted knowledge. Linguistic integrity, for its part, is insured
by the proposed method, and thus does not need to be evaluated.

The two criteria are actually independent of the OLS algorithm and
can be used to assess the quality of any Fuzzy Inference System
(FIS).

\subsection{Coverage index and activation threshold \label{covact}}

Consider a rule base $RB_r$ containing $r$ rules such as the one given in equation \ref{eq:rule}.

\textbf{Definitions}

Let $I_i$ be the interval corresponding to the \textit{ith} input range and $I^p \subseteq I_1 \times \ldots \times I_p$ be the subset of $\mathbb{R}^p$ covered by the rule base ($I_1 \times \ldots \times I_p$ is the Cartesian product).

\begin{definition}
An activation threshold $\alpha \in [0,1]$ defines the following
constraint:  \\
given $\alpha$, a sample $x^i$ is said active iff $\quad\exists \; R_q \in
RB_r$ s.t. $w^q(x^i) > \alpha$.
\end{definition}

\begin{definition}
Let $n$ be the number of active samples. The coverage index $CI_{\alpha}= n/N$ is the proportion of active samples for the activation threshold $\alpha$.
\end{definition}

Note:  Increasing the  activation threshold reduces the amount of active samples and transforms $I^p$ into a subset $I^p_{\alpha} \subseteq I^p$

The threshold choice  depends on the conjunctive operator used to compute the rule firing strength: the use of a \textbf{prod} operator yields lesser firing strengths which will decrease with the input dimension, while a \textbf{min} operator  results in higher and less dependent firing strengths. \\

The two-dimensional rule system depicted by figure~\ref{fig:covering} illustrates the usefulness
of the activation threshold and coverage index in the framework of knowledge extraction.

\subsubsection{Maximum coverage index ($\alpha=0$)}
$CI_0$ is the maximum coverage index and  it gives us two kinds of information:

\begin{figure}[htp]
\centering
\includegraphics[trim = 0mm 170mm 0mm 50mm, clip, scale=0.70]{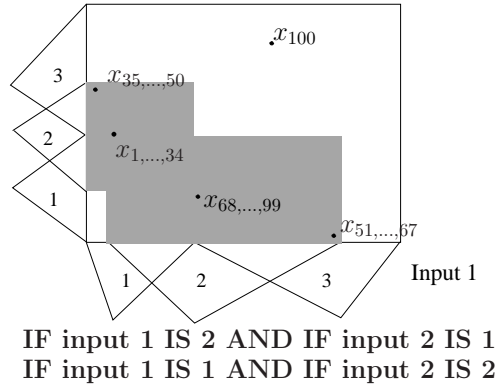}

\caption{Input domain rule coverage}
\label{fig:covering}
\end{figure}

\begin{itemize}
\item Completeness: for a so called complete system, where each data set item activates at least one rule, we have $CI_0=1$  while an empty rule base yields $CI_0=0$.
The coverage index can thus be used to measure the completeness of the rule base, with respect to a given data set.
\item  Exception data:  $CI_{\alpha}\approx 1$ is often the consequence of exception samples. We call exception an
isolated sample which is not covered by the rule base. Sample $x_{100}$ in figure~\ref{fig:covering} is such an exception.
\end{itemize}

The maximum coverage index of the system shown in figure~\ref{fig:covering}  is $99 \%$, meaning that there are a few exceptions.

\subsubsection{Coverage index for $\alpha>0$}

\begin{figure}[htp]
\centering
\includegraphics[trim = 0mm 170mm 0mm 50mm, clip, scale=0.70]{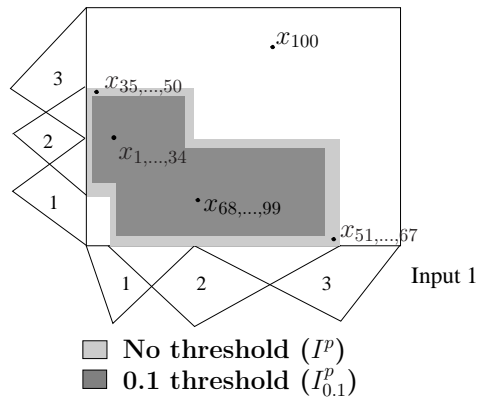}

\caption{Input domain with $\alpha=0.1$}
\label{fig:threshcovering}
\end{figure}

Figure~\ref{fig:threshcovering} shows the previous system behaviour with an activation threshold $\alpha = 0.1$. Unfortunately, the coverage index drastically drops from
$99 \%$ to $66 \%$.

Generally speaking, the use of a coverage index gives indications as to:
\begin{itemize}
\item System robustness (see Figure~\ref{fig:threshcovering}).

\item Reliability of extracted knowledge: Figure~\ref{fig:covering} shows that a
system can have a good accuracy and be unreliable.

\item Rule side effect: if too many samples are found in the rule borders, the rule base reliability is questionable.
Studying the evolution of coverage versus the activation threshold allows to quantify this "rule side effect".

\end{itemize}

As we shall see in section \ref{section:benchmark}, a blind application of OLS may induce fairly pathological situations (good accuracy and a perfect  coverage index dropping down as soon as an activation threshold constraint is added).

The coverage index and activation threshold are easy-to-use, easy-to-understand tools with the ability to detect such undesirable rule bases.

\subsection{Performance index}

The performance index reflects the numerical accuracy of the predictive system. In this study, we  use  $\displaystyle PI = \frac{1}{n} \sqrt {\sum_{i = 1}^n\left\| \widehat{y^i} - y^i \right\|^2}$.

The performance index only takes account of the active samples ($n\le N$, see definition 2), so a given system may have good prediction results on only a few of the available samples (i.e. good $PI$   but poor $CI{\alpha}$), or cover the whole data set, but with a lower accuracy.

\section{Proposed modifications for the OLS}\label{section:modifs}
In this section, we propose changes that aim to improve induced rule
interpretability. Rule premises, through variable partitioning, and
rule conclusions are both subject to modification.
\begin{figure}[htbp]
\begin{center}
\includegraphics[scale=1.2]{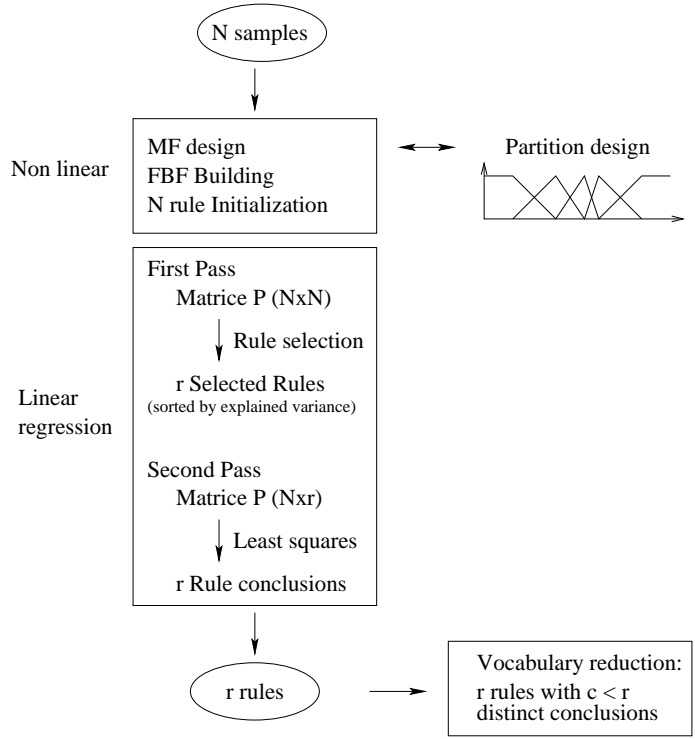}
\end{center}
\caption{Flowchart for the modified OLS algorithm}
\label{fig:flowchart}
\end{figure}

Figure \ref{fig:flowchart} is a flowchart describing the method used in the original OLS and the proposed modifications.

\begin{figure}[htbp]
\begin{center}
\includegraphics[scale=0.40]{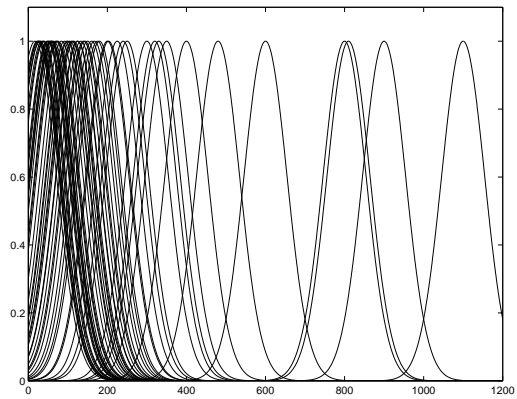}
\end{center}
\caption{Original fuzzy partition generated from a 106-item sample}
\label{fig:gauss1}
\end{figure}

The  fuzzy partitioning readability is a prerequisite to
build an interpretable rule base \cite{Guillaume01b}. In the original OLS algorithm, a rule is
built from each item of the training set, and a Gaussian membership
function is generated from each value of each variable. Thus a given fuzzy
partition is made up of as many fuzzy sets as there are distinct values within the
data distribution. The result, illustrated in figure \ref{fig:gauss1},
is not interpretable. Some membership functions are quasi redundant, and many of the corresponding fuzzy sets are not distinguishable, which makes it impossible to give them a semantic meaning.
Moreover, Gaussian functions have another drawback for our purpose: their unlimited boundaries, which yield a perfect coverage index, likely to drop down as soon as an activation threshold is set.

\subsection{Fuzzy partition readability}
The necessary conditions for a fuzzy partition to be interpretable by
a human expert have been studied by several authors
\cite{Valente99,Glorennec99,Espinosa00}. Let us recall the main
points:

\begin{itemize}
\item Distinguishability:  Semantic integrity requires that the membership functions represent linguistic concepts different from each other.
\item A justifiable number of fuzzy sets \cite{Miller}.
\item Normalization: All the fuzzy sets should be normal.
\item Overlapping: All the fuzzy sets should significantly overlap.
\item Domain coverage:  Each data point, $x$, should belong significantly, \mbox{$\mu(x) > \epsilon$}, at least to one fuzzy set. $\epsilon$ is called the coverage level \cite{Pedrycz93}.
\end{itemize}

We implement these constraints within a standardized fuzzy partition
as proposed in \cite{Ruspini69}:
\begin{equation}
\left\lbrace
\begin{array}{l}
\forall x \sum\limits_{f=1,2,\ldots,M} \ \mu_{f}(x)=1 \\
\forall f \ \exists \ x \ such \ as \ \mu_{f}(x)=1
\end{array}\right.
\label{pff}
\end{equation}
where $M$ is the number of fuzzy sets in the partition and $\mu_{f}(x)$ is the membership degree of $x$ to the $fth$ fuzzy set.
Equation \ref{pff} means that any point belongs at most to two fuzzy sets when the fuzzy sets are convex.

\begin{figure}[htbp]
\begin{tabular}{cc}
\includegraphics[scale=0.4]{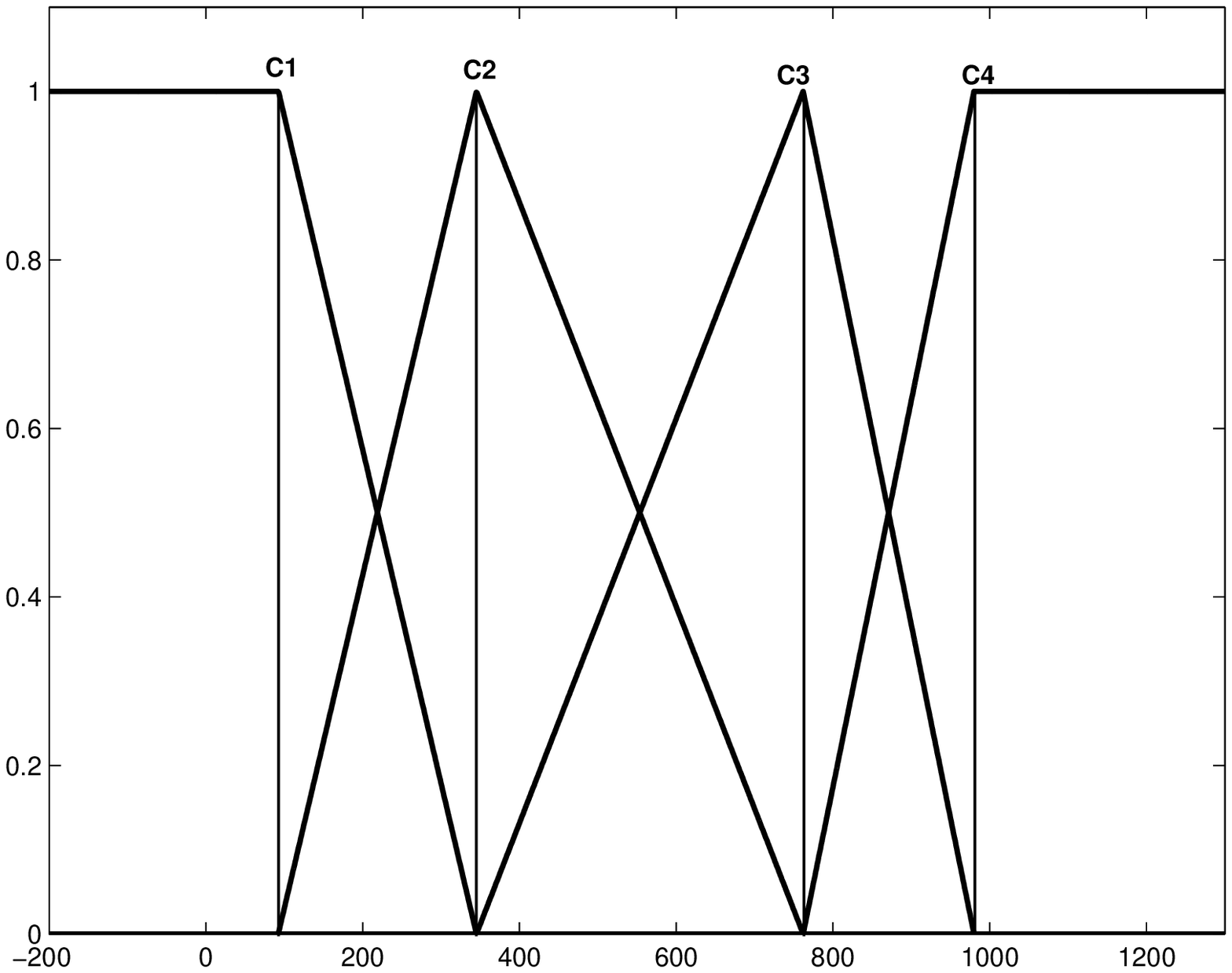} & \includegraphics[scale=0.4]{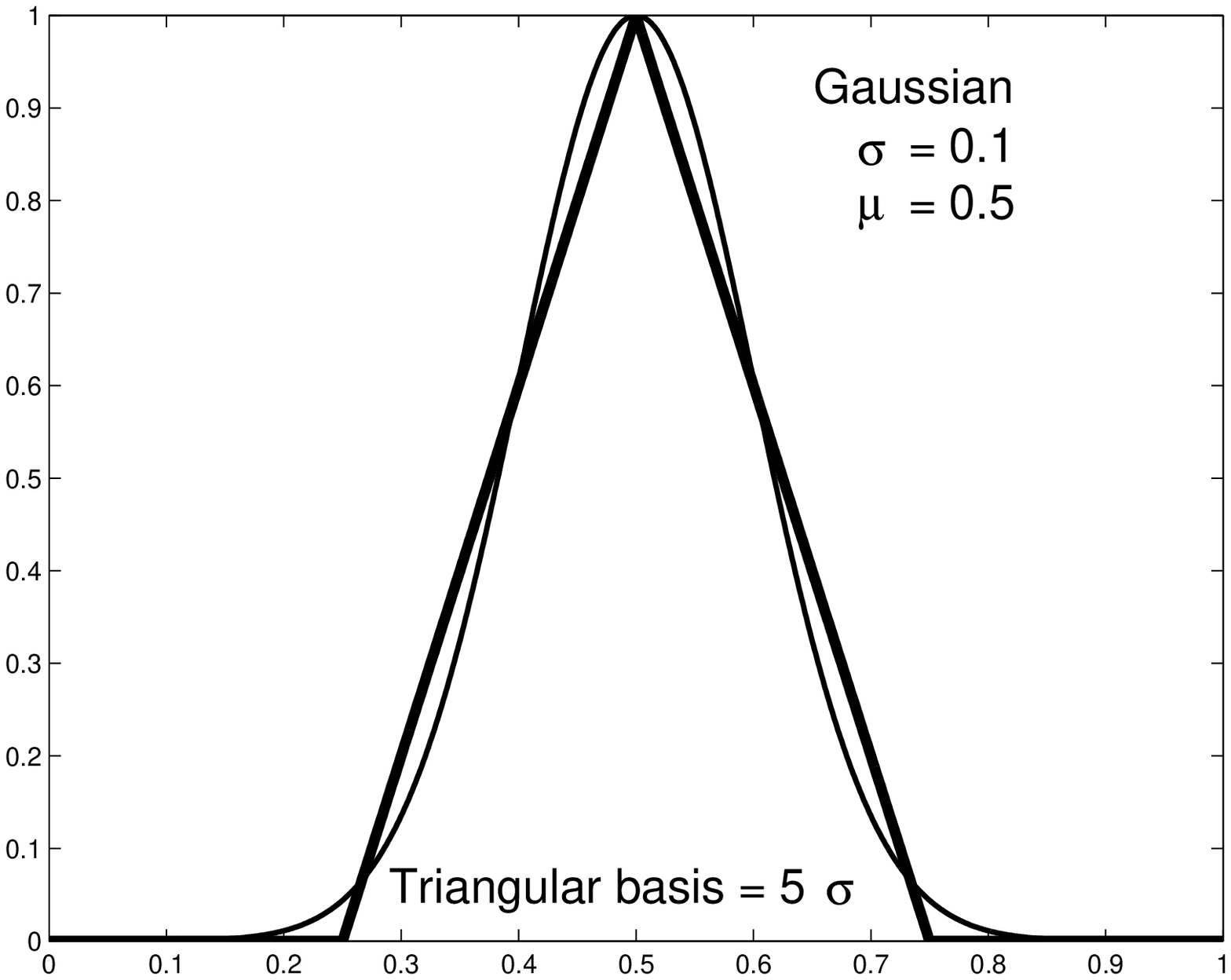}\\
 (a) A standardized fuzzy partition & (b) Triangle equivalent to a Gaussian MF\\
\end{tabular}
\caption{New fuzzy partitions}
\label{fig:strong}
\end{figure}

Due to their specific properties \cite{Pedrycz94} we choose fuzzy sets of triangular shape, except at the domain edges, where they
are semi trapezoidal, as shown in figure \ref{fig:strong}.a.
Such a $M$-term standardized fuzzy partition is completely defined by $M$ points, the fuzzy set centers.
With an appropriate choice of parameters, symmetrical triangle MFs approximately cover the same range as their Gaussian equivalent (see figure  \ref{fig:strong}.b).

\subsection{Fuzzy partition design\label{sec:fpdesign}}
Various methods are available to build fuzzy partitions \cite{Krishnapuram98}.
 In this paper, we want to use the OLS algorithm to build interpretable rule bases, while preserving a good numerical accuracy. To be sure that this is the case, we have to compare results obtained with the original partitioning design in OLS and those achieved with an interpretable partitioning.
To that effect, we need a simple and efficient way to design standardized fuzzy partitions from data, as the one given below.
The fuzzy set centers are not equidistant as in a regular grid, but are estimated according to the data distribution, using the well known $k-means$ algorithm \cite{Hartigan79}. The multidimensional k-means is recalled in Algorithm \ref{algokmeans}, we use it here independently in each input dimension.
\begin{algorithm}

\begin{algorithmic}[1]

\STATE Let N multidimensional data points denoted $x_i, i=1 \ldots N$

Let C the number of clusters to build
\STATE Initialization: choose k centroids E(k), $k=1\ldots C$

(random or uniformly spaced)

\STATE Assign each data point to the nearest cluster:

$cluster(x_i)=argmin_{k}(dist(x_i,E(k))$
\STATE Compute cluster centroids m(k)
\WHILE{$\exists k \quad \mbox{such as} \quad m(k) \neq E(k)$}

\STATE FOR (k=1 to C) $E(k)=m(k)$
\STATE GOTO 3

\ENDWHILE

\end{algorithmic}
\caption{k-means algorithm}
\label{algokmeans}
\end{algorithm}

How to choose the number of fuzzy sets for each input variable?
There are several criteria to assess partition quality \cite{Guillaume01b} but it is difficult to make an a priori choice.
 In order to choose the appropriate partition size, we first generate a hierarchy of partitions of increasing size in each input dimension $j$, denoted $FP_j^{n_j}$ for a $n_j$ size, $n_j^{max}$ being the maximum
size of the partition (limited to a reasonable number ($\approx 7$) \cite{Miller}).

Note:  $FP_j^{n_j}$ is uniquely determined by its size $n_j$, the fuzzy set
centers being the coordinates computed by the $k-means$ algorithm.

The best suited number of terms for each input variable is determined
using a refinement procedure based on the use of the hierarchy of fuzzy
partitions. This iterative algorithm is presented below. It calls a FIS generation
algorithm to be described later. It is not a greedy algorithm, unlike
other techniques. It does not implement all possible combinations of
the fuzzy sets, but only a few chosen ones.

\begin{table}
\begin{tabular}{l|l|llll|ll}
Line \# & Iteration \# & \multicolumn{4}{c|}{\#MF per variable}& PI & CI\\
\hline
1 & 1 & 1 & 1 & 1 & 1 & $PI_1$ & $CI_1$\\
\hline
\hline
2 & 2 & 2 & 1 & 1 & 1 & $PI_2^1$ & $CI_2^1$\\
3 & 2 & 1 & 2 & 1 & 1 & $PI_2^2$ & $CI_2^2$(best)\\
4 & 2 & 1 & 1 & 2 & 1 & $PI_2^3$ & $CI_2^3$\\
5 & 2 & 1 & 1 & 1 & 2 & $PI_2^4$ & $CI_2^4$\\
\hline
\hline
6 & 3 & 2 & 2 & 1 & 1 & $PI_3^1$ & $CI_3^1$\\
7 & 3 & 1 & 3 & 1 & 1 & $PI_3^2$ & $CI_3^2$\\
8 & 3 & 1 & 2 & 2 & 1 & $PI_3^3$ & $CI_3^3$ (best)\\
9 & 3 & 1 & 2 & 1 & 2 & $PI_3^4$ & $CI_3^4$\\
\hline
\hline
10 & 4 & 2 & 2 & 2 & 1 & $PI_4^1$ & $CI_4^1$\\
11 & 4 & \multicolumn{4}{c|}{\ldots}& \multicolumn{2}{c}{\ldots}\\
\end{tabular}
\caption{An example of ongoing refinement procedure}
\label{tab:refinement}
\end{table}

Table \ref{tab:refinement} illustrates the first steps of a refinement procedure for a four inpur system. Detailed procedures are  given in Algorithm \ref{algorefine}(refinement procedure) and Algorithm \ref{algogene} (FIS generation).
\begin{algorithm}
\begin{algorithmic}[1]


\STATE  iter = 1; $\forall j \  n_j=1$
\STATE  CALL \textit{FIS Generation} (Algorithm \ref{algogene})
\WHILE {$iter \le iter_{max}$}
\STATE  Store system as base system

\FOR{$1 \le j \le p $}
\STATE {\bf if} {$n_j = n_j^{max}$} {\bf then} next j (partition size limit reached for input j)
\STATE $n_j$ = $n_j$ + 1
\STATE CALL \textit{FIS Generation} (Algorithm \ref{algogene})
\STATE  $PI_j = PI$
\STATE $n_j$ = $n_j$ - 1
\STATE  Restore base system
\ENDFOR

\STATE {\bf if} $\forall j \ n_j = n_j^{max}  $ {\bf then} exit (no more inputs to refine)


\STATE s = argmin $\{PI_j, \ j=1,\ldots,p, \quad n_j <  n_j^{max} \}$ (Select input to refine)
\STATE $n_s$ = $n_s$ + 1
\STATE CALL \textit{FIS Generation} (Algorithm \ref{algogene}):  return $FIS_{iter}$
\STATE iter = iter + 1
\ENDWHILE

\end{algorithmic}

\caption{Refinement procedure}
\label{algorefine}
\end{algorithm}

The key idea is to introduce as many variables, described by a sufficient number of fuzzy sets, as necessary to get a good rule base.

The initial FIS is the simplest one possible, having only one rule (Algorithm \ref{algorefine}, lines 1-2; Table \ref{tab:refinement}, line 1). The search loop (algorithm lines 5 to 12) builds up temporary fuzzy inference systems. Each of them corresponds to adding to the initial FIS one fuzzy set in a given dimension. The selection of the dimension to retain is based upon performance and is done in lines 14-15 of the algorithm. If we go back to table \ref{tab:refinement}, we  see that the second iteration corresponds to lines 2 to 5, and that the best configuration is found by refining input variable \# 2.
Following this selection, a FIS to be kept is built up. It will serve as a base to reiterate the sequence (Algorithm \ref{algorefine}, lines 3 to 18).

When necessary, the procedure calls a FIS generation algorithm, referred to as Algorithm \ref{algogene}, which is now detailed.

The rule generation is done by combining the fuzzy sets of the $FP_j^{n_j}$ partitions for $j=1,\ldots,p$, as described by Algorithm \ref{algogene}. The algorithm then removes the less influential rules and evaluates the rule conclusions, using output training data values $y^i, i=1\ldots N$.
\begin{algorithm}

\begin{algorithmic}[1]

\REQUIRE $\{n_j \ | \ j = 1,...,p\}$

\STATE get $FP_j^{n_j} \quad \forall j=1,...,p$\\

\STATE Generate the $\prod_{j=1}^p n_j$ rule premises
\FORALL{Rule $r \in FIS$}
        \STATE{ $\alpha_r = \max \limits_{k} w^r(x^k)$}
            \STATE {\bf if} $\alpha_r <\alpha$ {\bf then} remove rule $r$
        \STATE {\bf else} initialize rule conclusion $C_r=\frac{1}{N}\sum\limits_{i=1}^N w^r y^i$
\ENDFOR


\STATE  Compute PI
\end{algorithmic}
\caption{FIS generation}
\label{algogene}
\end{algorithm}
The condition stated in line $5$, where $\alpha$, the activation threshold defined in section \ref{covact},
ensures that the rule is significantly fired by the examples of the
training set.

The procedure does not yield a single fuzzy inference system,
but $K$ FIS of increasing complexity.  The selection of the best one
takes into consideration both performance and  coverage indices.
The selected $FIS_k$ corresponds to:

$k=argmin(PI_k, k=1,\ldots,K ~ such ~  as ~  CI_{\alpha}(FIS_k) \ge thres$), where
$PI_k$ and $CI_{\alpha}(FIS_k)$ are the $FIS_k$ performance and coverage indices.

In the following, only the fuzzy partition corresponding to the \textit{best} FIS will be kept. The initial rules are ignored as they will be determined by the OLS.

The use of standardized fuzzy partitions, with a small number of
linguistic terms, ensures that rule premises are interpretable. Moreover, that choice eliminates the problem of \textit {quasi redundant} rule selection, due to MF redundancy and underlined by authors familiar with these procedures, as \cite{Setnes03}.

However, the OLS brings forth a different conclusion for each rule. It makes rules difficult to interpret. We will now propose another modification of the OLS procedure to improve that point.

\subsection{Rule conclusions}\label{section:redvoc}

Reducing the number of distinct output values improves
interpretability as it makes rule comparison easier. Rule conclusions may be assigned a linguistic
label if the number of distinct conclusions is small enough.
The easiest way to reduce the number of distinct output values is to adjust conclusions upon completion of the algorithm. We use the following method based on the k-means algorithm.

Given a number of distinct conclusions, $c$, and the set of training
output values $y^i, \ i=1,2,\ldots,N$, the reduction procedure consists in:

\begin{itemize}
\item Applying the k-means method \cite{Hartigan79} to the $N$ output values with $c$ final clusters.
\item For each rule of the rule base, replacing the original conclusion by the nearest one obtained from the k-means.
\end{itemize}

The vocabulary reduction worsens the system numerical
accuracy on training data. However, rule conclusions are no more computed only with a least square optimization, and the gap between training and test errors may be reduced.

\section{Results on benchmark data sets}\label{section:benchmark}
This section presents, compares and discusses results obtained on two well known cases chosen in the UCI repository \cite{UCI}.

\subsection{Data sets}\label{datasets}
The data sets are the following ones:
\begin{itemize}
\item \textit{cpu-performance} (209 samples): \\ Published by Ein-Dor and Feldmesser \cite{Ein-Dor}, this data set contains the measured CPU performance and 6 continuous variables such as main memory size or machine cycle time.
\item \textit{auto-mpg} (392 samples): \\ Coming from the StatLib
library maintained at Carnegie Mellon University, this case concerns
the prediction of city-cycle fuel consumption in miles per gallon from
4 continuous and 3 multi-valued discrete variables.

\end{itemize}
The cpu-performance and auto-mpg datasets are both regression problems. Many results have been reported for them in the previous years \cite{Quinlan93, Zurada04, Pedrycz, Setiono}.

\vspace{0.05cm}
\textbf{Experimental method}

For the experiments, we use on each dataset a ten-fold cross validation method. The entire dataset is randomly divided into ten parts. For each part, the training is done on the nine others while testing is made on the selected one. Besides the stop criterion based upon the cumulated explained variance (equation \ref{eq:thresh}), another one is implemented: the maximum number of selected rules. The algorithm stops whenever any of them is satisfied.

\subsection{Results and discussion}

Tables \ref{tab:cpu} and \ref{tab:auto}  summarize the results for original and modified OLS on test subsets, for each data set. The original OLS algorithm is applied with only a slight modification: the conjunction operator in rule premises is the minimum operator instead of the product. Tests have been carried out to check that results are not significantly sensitive to the choice of the conjunction operator. The choice of the minimum allows a fair comparison between data sets  with a different number of input variables.

Both tables have the same structure: the first column gives the average number of membership functions per input variable, the following ones are grouped by three.
Each group of three corresponds to a different value of the allowed maximum number of rules, ranging from unlimited to five.
The first one of the three columns within each group is the average number of rules, the second one the average performance index $PI$ and the third one is the average coverage index $CI_{\alpha}$, which corresponds to the activation threshold $\alpha$ given in the row label between parentheses.
The first group of three columns corresponds to an unlimited number of rules, the actual one  found by the algorithm being given in the \#R column.

\setlength{\tabcolsep}{1.7pt}
\renewcommand\arraystretch{0.9}
\begin{small} \begin{center}
\begin{table}[htbp]

\begin{tabular}{|l|r|lll|lll|lll|lll|}
\hline

& \#MF & \#R & Perf. & Cov. & \#R & Perf. & Cov. & \#R & Perf. & Cov. & \#R & Perf. & Cov. \\
\hline
orig. OLS (0) & 27.8 & 39.8 & 69.78 & 1.00 & 15 & 74.54 & 1.00 & 10 & 98.11 & 1.00 & 5 & 150.38 & 1.00 \\
orig. OLS (0.1) & 27.8 & 39.8 & 32.52 & 0.75 & 15 & 33.32 & 0.40 & 10 & 46.65 & 0.23 & 5 & 113.26 & 0.03 \\
orig. OLS (0.2) & 27.8 & 39.8 & 32.30 & 0.59 & 15 & 40.67 & 0.21 & 10 & 61.99 & 0.09 & 5 & 113.26 & 0.03 \\
\hline
mod. OLS (0) & 2.7 & 11.3 & 41.95 & 0.99 & 11.3 & 41.95 & 0.99 & 10 & 45.57 & 0.97 & 5 & 71.96 & 0.47 \\
mod. OLS (0.1) & 2.7 & 11.3 & 41.95 & 0.99 & 11.3 & 41.95 & 0.99 & 10 & 46.01 & 0.95 & 5 & 71.71 & 0.45 \\
mod. OLS (0.2) & 2.7 & 11.3 & 41.92 & 0.98 & 11.3 & 41.92 & 0.98 & 10 & 45.07 & 0.91 & 5 & 69.27 & 0.40 \\
\hline
\end{tabular}
\caption{Cpu data comparison of original and modified OLS (averaged on 10 runs)}
\label{tab:cpu}

 \end{table}\end{center}
\end{small}

\begin{small}
\begin{center}
\begin{table}[htbp]

\begin{tabular}{|l|r|lll|lll|lll|lll|lll|}

\hline
 & \#MF & \#R & Perf. & Cov. & \#R & Perf. & Cov. & \#R & Perf. & Cov. & \#R & Perf. & Cov. & \#R & Perf. & Cov. \\
\hline

orig. OLS (0) & 86.8 & 182.9 & 3.31 & 1.00 & 20 & 3.88 & 1.00 & 15 & 4.32 & 1.00 & 10 & 5.47 & 1.00 & 5 & 9.35 & 1.00 \\
orig. OLS (0.1) & 86.8 & 182.9 & 2.91 & 0.84 & 20 & 3.08 & 0.40 & 15 & 3.22 & 0.34 & 10 & 3.35 & 0.25 & 5 & 3.55 & 0.18 \\
orig. OLS (0.2) & 86.8 & 182.9 & 2.75 & 0.77 & 20 & 2.92 & 0.32 & 15 & 3.11 & 0.27 & 10 & 3.21 & 0.21 & 5 & 3.22 & 0.15 \\
\hline
mod. OLS (0) & 3.3 & 19.3 & 3.03 & 1.00 & 19.3 & 3.03 & 1.00 & 15 & 3.05 & 1.00 & 10 & 2.99 & 0.99 & 5 & 3.33 & 0.90 \\
mod. OLS (0.1) & 3.3 & 19.3 & 3.03 & 1.00 & 19.3 & 3.03 & 1.00 & 15 & 3.05 & 1.00 & 10 & 2.99 & 0.99 & 5 & 3.36 & 0.85 \\
mod. OLS (0.2) & 3.3 & 19.3 & 3.03 & 1.00 & 19.3 & 3.03 & 1.00 & 15 & 3.05 & 1.00 & 10 & 3.00 & 0.98 & 5 & 3.36 & 0.81 \\
\hline

\end{tabular}

\caption{Auto-mpg data comparison of original and modified OLS (averaged on 10 runs)}
\label{tab:auto}
\end{table}
\end{center}
\end{small}

The discussion includes considerations about complexity, coverage and numerical accuracy of the resulting FIS.\\
Let us first comment the FIS structures. Clearly the original OLS yields a more complex system than the modified one, with a much higher number of membership functions per input variable.
When the number of rules is not limited, the original OLS systematically has many more rules than the modified one.
As to the performances, let us focus on rows one and four, which correspond to $\alpha=0$, and on the first three columns, to allow an unlimited number of rules. This configuration allows a fair comparison between both algorithms. We see that, for both data sets, the modified algorithm has an enhanced performance. For the cpu data set, this has a slight   coverage cost, with a loss of one percent, meaning that an average of two items in the data set is not managed by the systems obtained by the modified algorithm.\\
Examination of the next rows ($\alpha=0.1$) shows that the  modified algorithm systems have the same $PI$ and $CI_{\alpha}$ than for the zero threshold. It is not at all the case for the  original algorithm systems, where the coverage loss can be important (from 16 to 25 percent). This well demonstrates the lack of robustness of the original algorithm, as a slight change in input data may induce a significant output variation. The modified algorithm does not have this drawback.

\begin{figure}[htbp]
\begin{center}
\begin{tabular}{c}
\includegraphics[scale=0.4,angle=270]{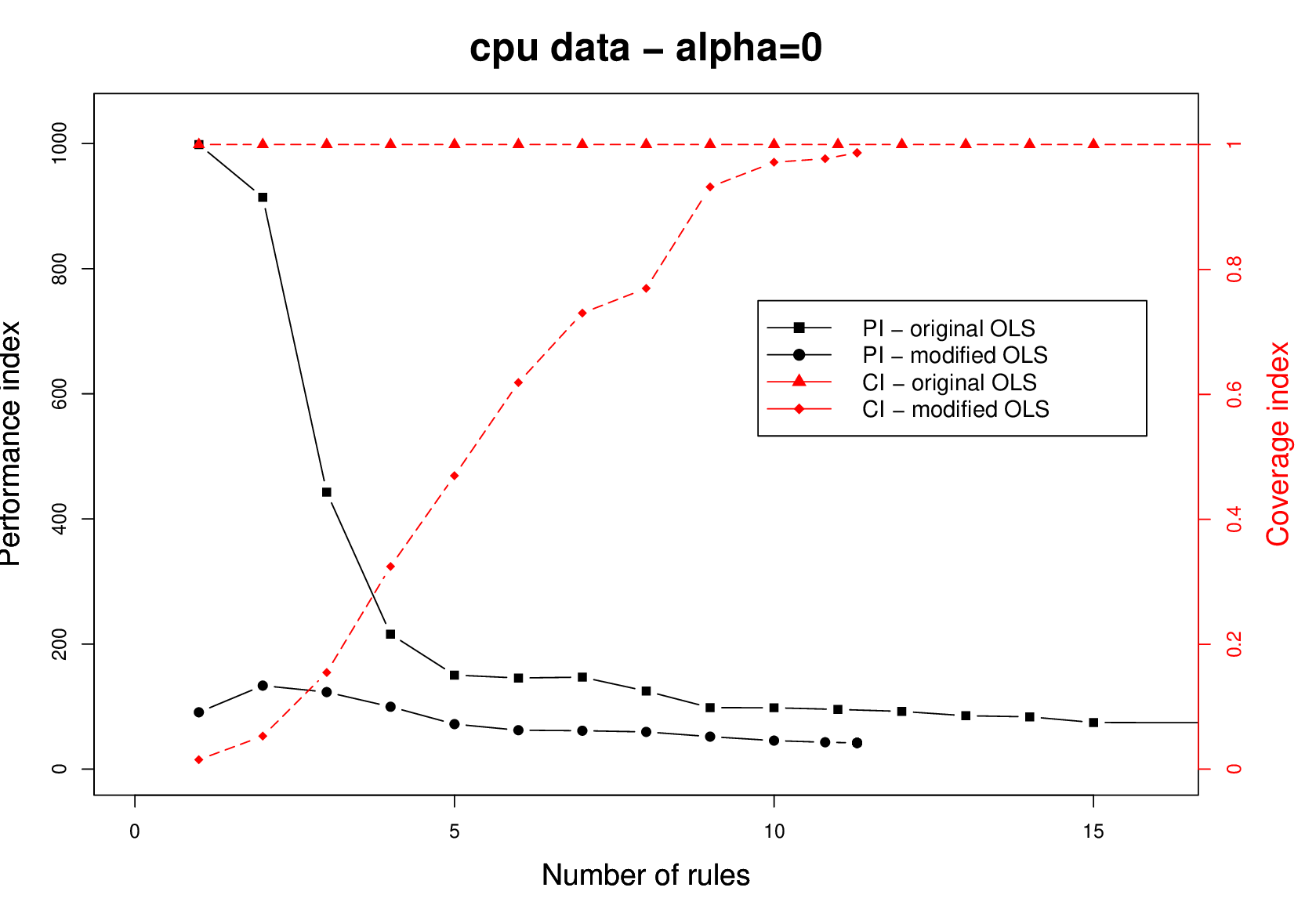}\\
\includegraphics[scale=0.4,angle=270]{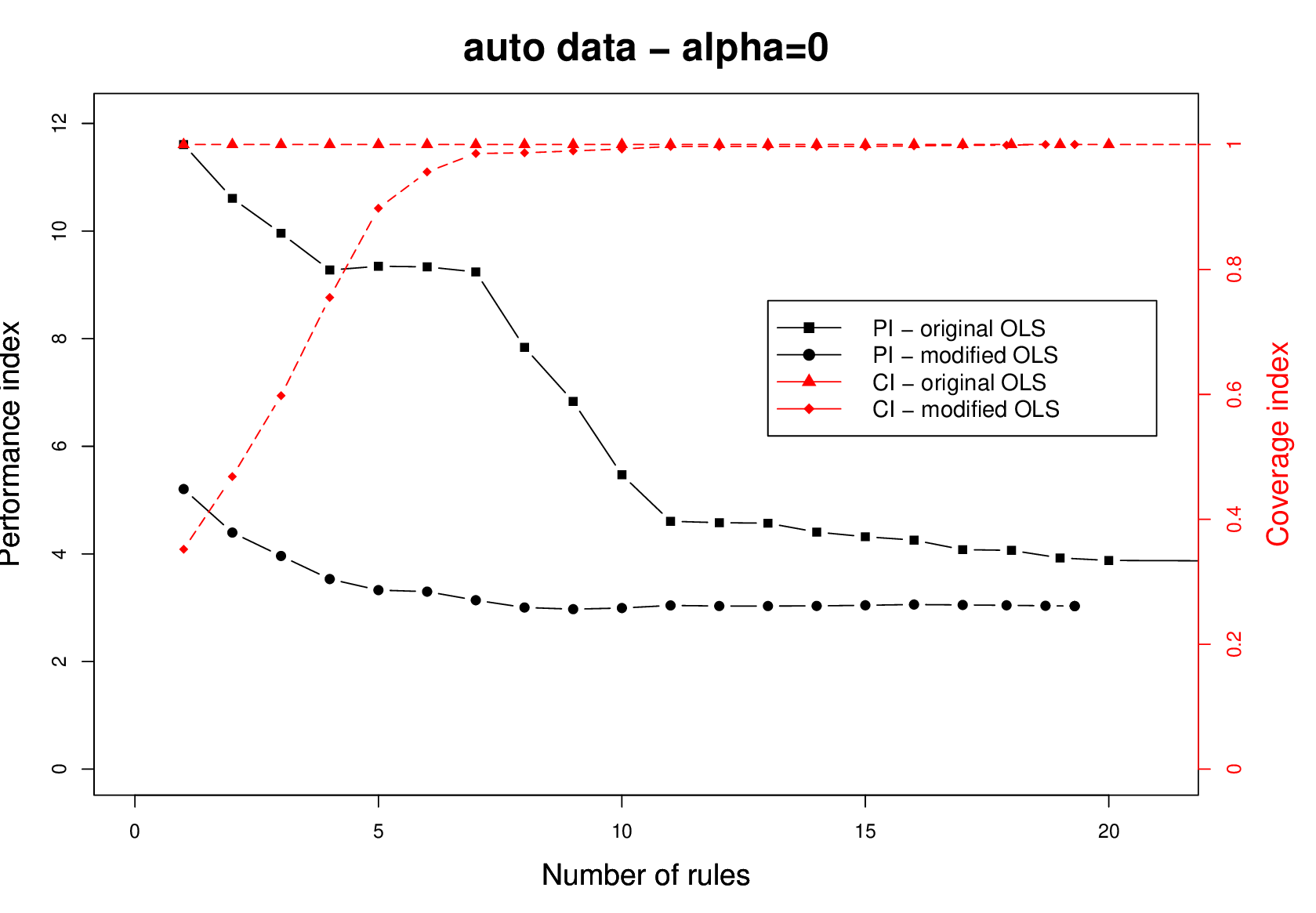}
\end{tabular}

\caption{Evolution of performance $PI$ and coverage index $CI_0$ versus number of rules}
\label{fig:evol}
\end{center}
\end{figure}

Figure \ref{fig:evol} shows the evolution of $CI_0$ and $PI$ with the number of rules for each data set.
As expected, the coverage index $CI_0$ is always equal to 1 for the original version.
For the modified version, $CI_0$ quasi linearly increases with the number of rules.
It means that each newly selected rule covers a set of data items, so that rules are likely to be used for knowledge induction, as will be shown in more details in section \ref{sec:lbe}.

For a reasonable number of rules ($\ge 10$), we see that, while $CI_0 \approx 1$, the modified OLS has a much better accuracy than the original one.\\
For a low number of rules, the performance index $PI$ has a very
different behaviour for the two OLS versions. The poor accuracy
(high values of $PI$) of the original algorithm can be explained by
a  low cumulated explained variance, and the good accuracy observed
for the modified algorithm must be put into perspective of its poor
$CI_0$. As the number of rules increases, both systems display a
similar behaviour.

Another advantage of the modified OLS noticed in the benchmark
results is the reduced execution time. When averaged over ten
runs with an unlimited number of selected rules, for the CPU and auto cases, it respectively took 1.16 s and
5.65 s CPU on a 32 bit Xeon 3.2 GHz processor for the original OLS algorithm to complete, while it
respectively took 1.03 s and 3.72 s  for the modified version.

Table \ref{tab:compare} compares the results of the modified OLS method and of other methods used in the literature (see \cite{Quinlan93}),  in terms of Mean Absolute Error (criterion used in that reference paper), computed as $MAE=\frac{1}{n} \sum\limits_{i=1}^{n} |\widehat{y}_i - y_i |$, $n$ being the number of active samples.
The first method is a multivariate linear regression (LR), the second one is a regression tree (RT) and the third one is a neural network (NN).
In all cases, the modified OLS average error is comparable to those of competing  methods, or even better.

\begin{table}[htbp]

\begin{center}

\begin{tabular}{|c|c|c|c|c|}
\hline
Data set & Mod.OLS & Linear regression & Decision tree & Neural network \\ \hline
\textit{CPU-Performance} & 28.6 & 35.5 & 28.9 & 28.7 \\
\textit{Auto-mpg} & 2.02 & 2.61 & 2.11 & 2.02 \\ \hline
\end{tabular}
\end{center}
\caption{Comparison of mean absolute error of the modified OLS and other methods on test sets}
\label{tab:compare}
\end{table}

We showed in this section that the proposed modifications of the OLS algorithm yield good results on benchmark data sets. We will thus use the modified OLS to deal with a real world case.

\section{A real world problem}\label{sec:lbe}
The application concerns a fault diagnosis problem in a wastewater
anaerobic digestion process, where the "living" part of the biological process
must be monitored closely. Anaerobic digestion is a set of biological
processes taking place in the absence of oxygen and in which organic
matter is decomposed into biogas.

Anaerobic processes offer several advantages: capacity to treat
slowly highly concentrated substrates, low energy requirement and
use of renewable energy by methane combustion.
Nevertheless, the instability of anaerobic processes (and of the
attached microorganism population) is a counterpart that discourages
their industrial use. Increasing the robustness of such
processes and optimizing fault detection methods to efficiently
control them is essential to make them more attractive to
industrials. Moreover, anaerobic processes are in general very long
to start, and avoiding breakdowns has significant economic implications.

The process has different unstable states: hydraulic
overload, organic overload, underload, toxic presence, acidogenic
state. The present study focuses on the acidogenic state. This state is
particularly critical, and going back to a normal state is time consuming,
thus it is important to detect it as soon as possible. It is mainly
characterized by a low pH value ($< 7$), a high concentration in
volatile fatty acid and a low alkalinity ratio (generally $< 0.3$).

Our data consist of a set of 589 samples coming from a pilot-scale
up-flow anaerobic fixed bed reactor (volume=0.984 $m^3$). Data are provided by the LBE, a laboratory situated
in Narbonne, France. Seven input variables summarized in table~\ref{tab:inpvar} were used in the case study.

\begin{table}
  \centering
  \begin{tabular}{|c|c|}
    \hline
    Name & Description \\
    \hline
    $pH$ & pH in the reactor \\
    $vfa$ & volatile fatty acid conc. \\
    $qGas$ & biogas flow rate \\
    $qIn$ & input flow rate \\
    $ratio$ & alkalinity ratio \\
    $CH_4Gas$ & $CH_4$ concentration in biogas \\
    $qCO_2$ & $CO_2$ flow rate \\
    \hline
  \end{tabular}
  \caption{Input variables}\label{tab:inpvar}
\end{table}

The output is an expert assigned  number from 0 to 1 measuring to what extent the actual state can be
considered as acidogenic.

Fault detection systems in bioprocesses are usually based on expert knowledge. Multidimensional interactions are imperfectly known by experts. The OLS method allows to build a fuzzy rule base from data, and the rule induction can help experts to refine their knowledge of fault-generating process states.

\begin{figure}

\begin{center}
\begin{tabular}{cc}
\includegraphics[width=0.45\textwidth]{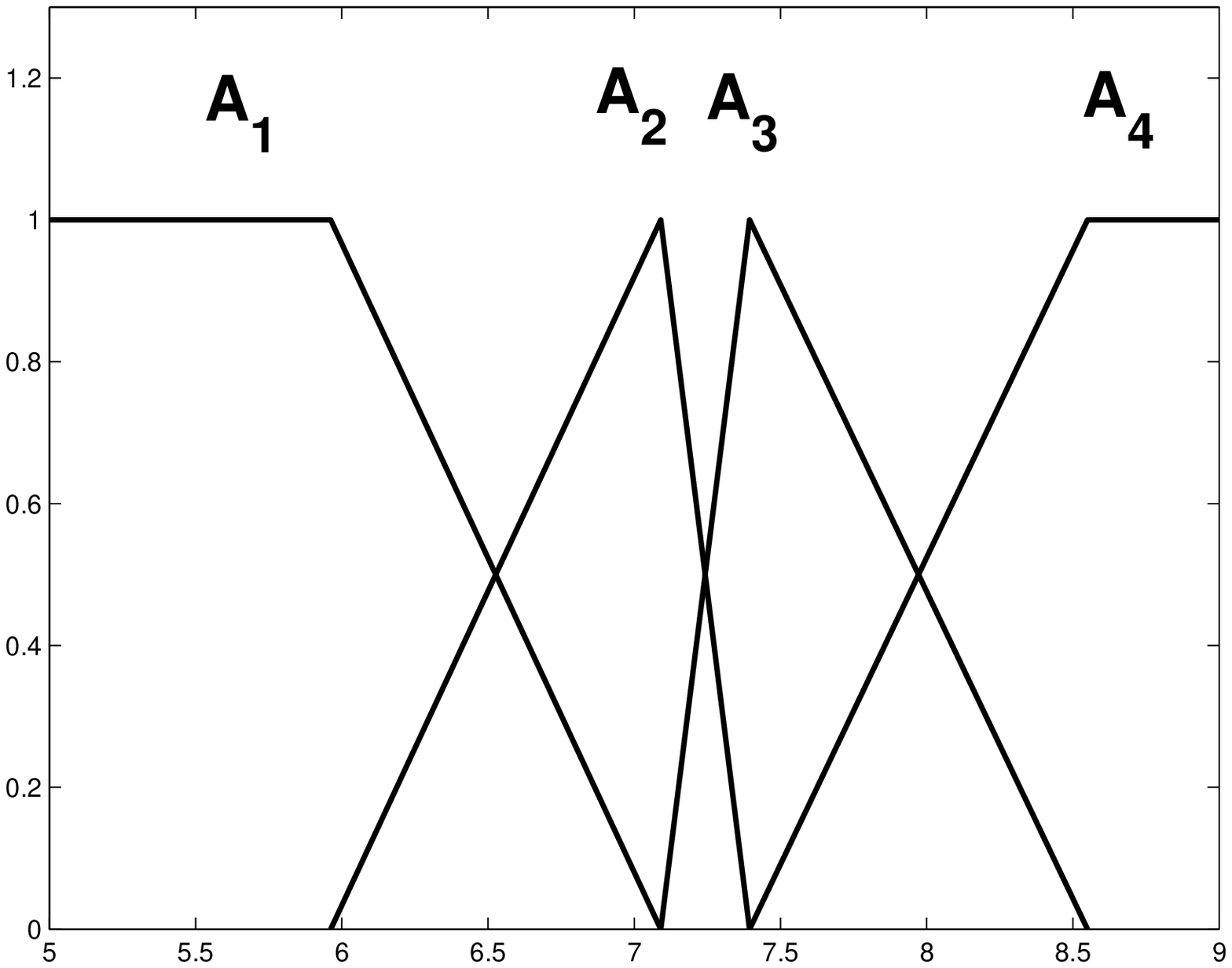} & \includegraphics[width=0.45\textwidth]{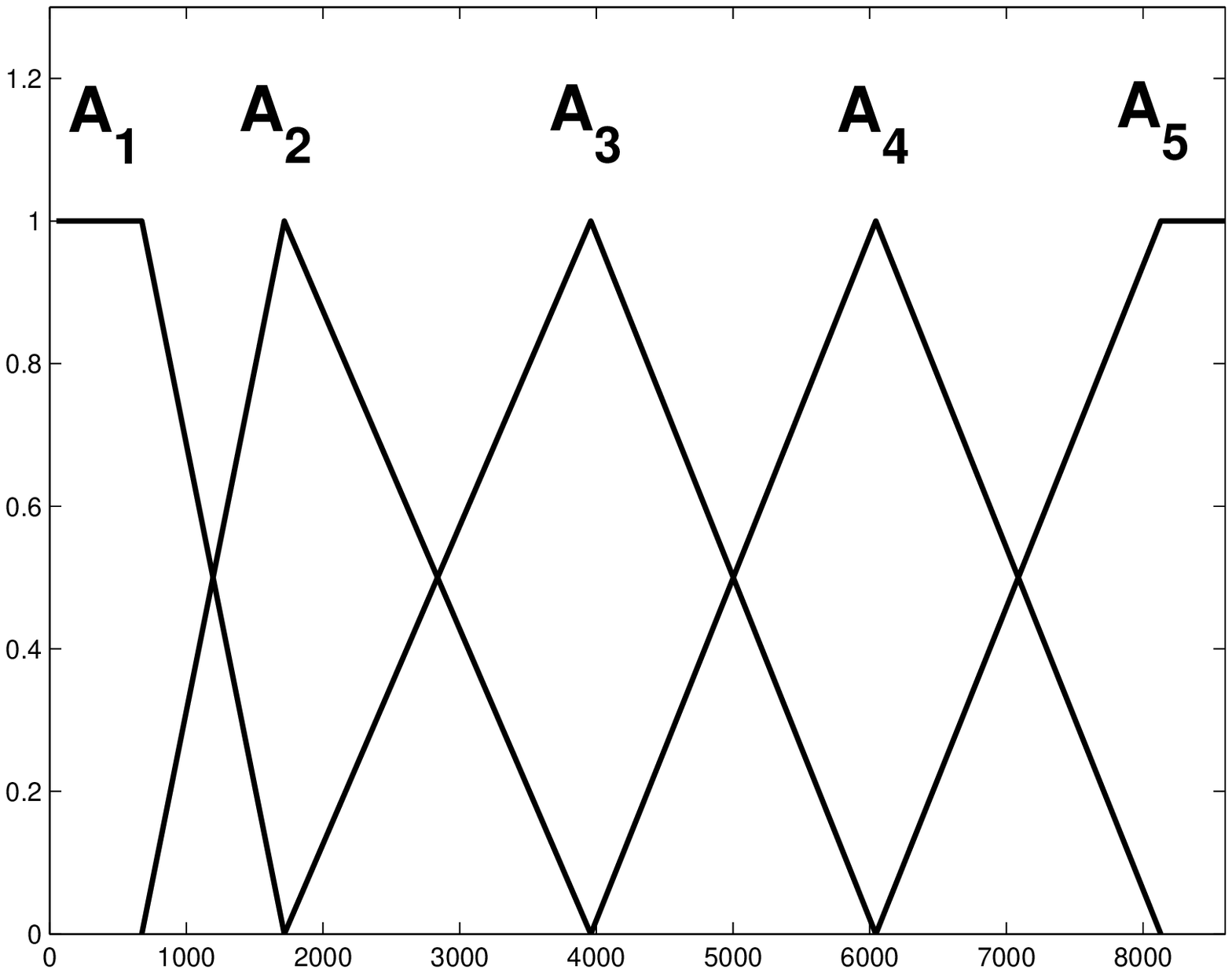} \\
pH & vfa \\
\includegraphics[width=0.45\textwidth]{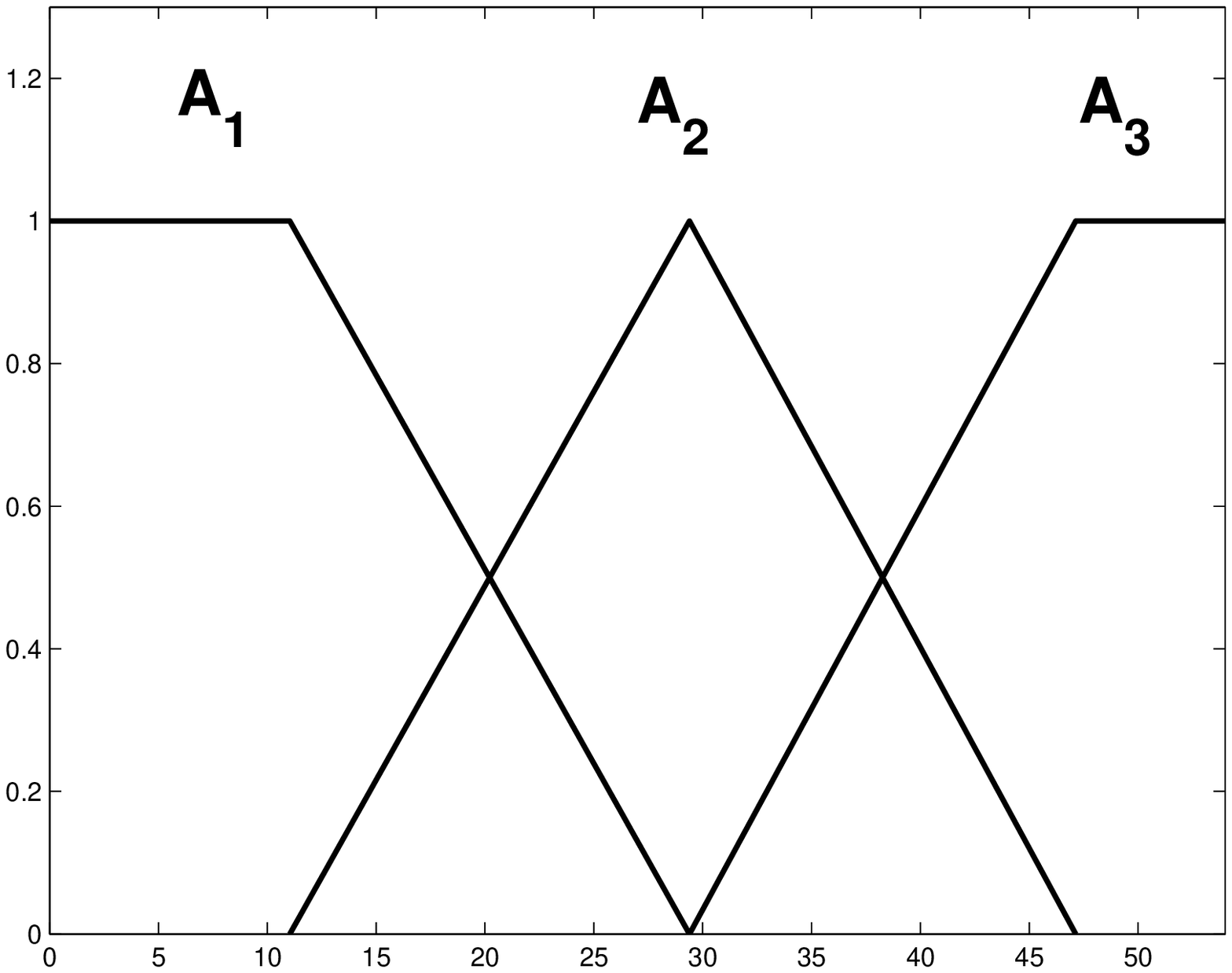} &
\includegraphics[width=0.45\textwidth]{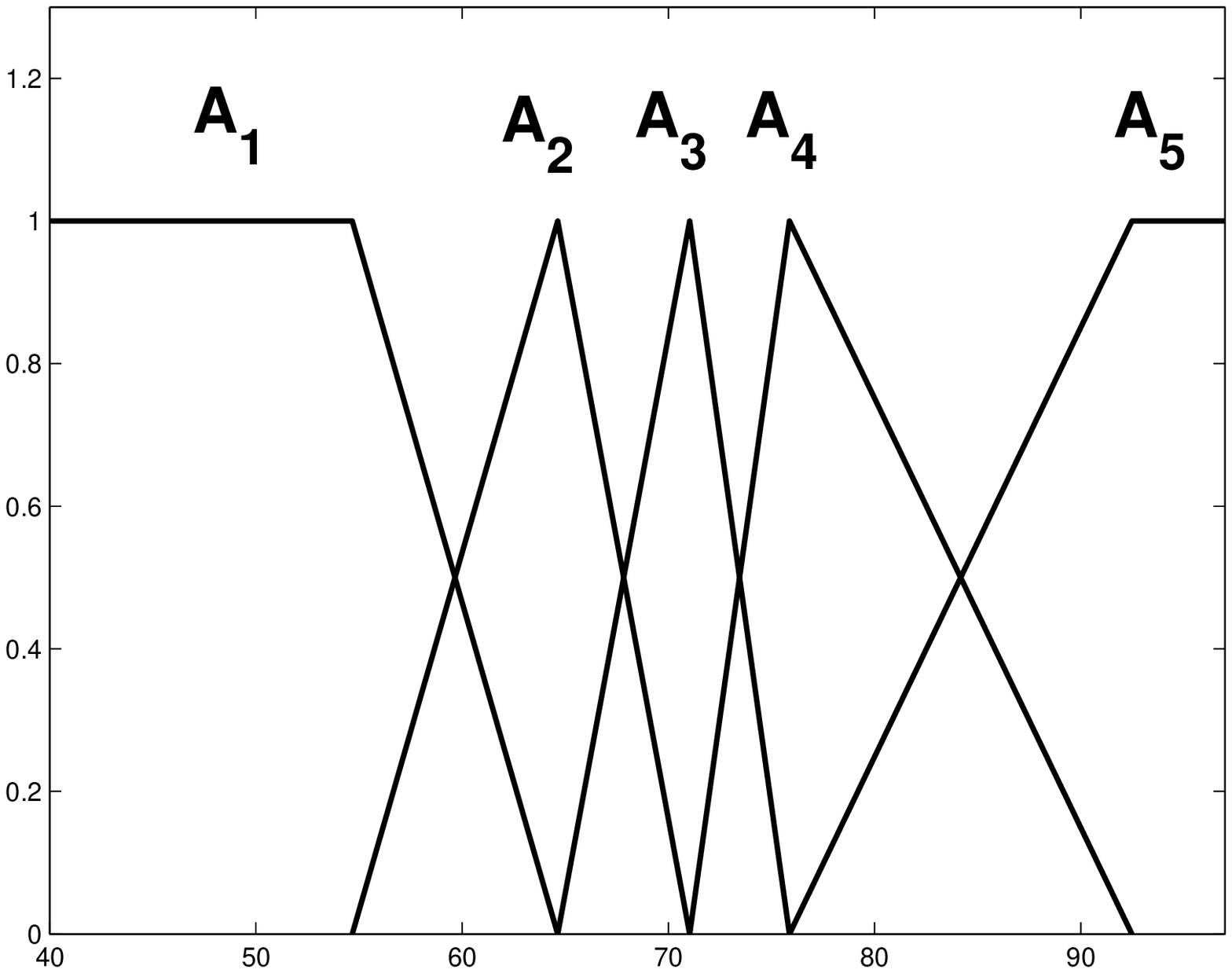}
\\
Input flow rate ($Qin$) & $CH_4$
\end{tabular}
\end{center}
\caption{Fuzzy partitions for wastewater treatment application}
\label{fig:fuzzpart}
\end{figure}

Before applying the OLS, we select the fuzzy partition with the refinement algorithm described in section \ref{section:modifs}, which yields the selection of  four input variables : $pH$, $vfa$, $Qin$ and
$CH_4Gas$.
The membership functions are shown in Figure~\ref{fig:fuzzpart}. Notice that each membership function can be assigned an interpretable linguistic label.

\textbf{Results and discussion}

We  apply the OLS procedure to the whole data set, obtaining a rule base
of 53 rules and a global performance $PI=0.046$.

\textbf{Rule base analysis}

Analyzing a rule base is usually a very long task, and must be
done anew with each different problem. Here are some general remarks:

\begin{itemize}
\item \textbf{Rule ordering}: amongst the 589 samples, only 35 have
an output value greater than 0.5 (less than $10\%$), while there are 12
rules out of 53 that have a conclusion greater than 0.5 (more than
$20\%$). Moreover, 8 of these rules are in the  first ten selected ones (the
first six having a conclusion very close to one). This shows
that the algorithm first select rules corresponding to "faulty"
situations. It can be explained by the fact that the aim of the
algorithm is to reduce variance, a variance greatly increased by a
"faulty" sample. This highlights a very interesting characteristic
of the OLS algorithm, which first selects rules related to rare
samples, often present in fault diagnosis.

\item \textbf{Out of range conclusions}: each output in the data set
is between 0 and 1. This is no more the case with the rule
conclusions, some of them being greater than 1 or taking negative values.
It is due to the least-square optimization method trying to
improve the accuracy by adjusting rule conclusions, without
any constraint. This is one of the deficiencies of the
algorithm, at least from an interpretability driven point of view.

\end{itemize}

\textbf{Removing outliers}

The fact that rules corresponding to rare samples are favored in the
selection process has another advantage: the ease with which
outliers can be identified and analyzed. In our first rough analysis
of the rule base, two specific rules caught our attention :

\begin{itemize}
\item \textbf{Rule 5} : If pH is $A_3$ and vfa is $A_1$ and Qin is
$A_1$ and $CH_4$ is $A_1$, then output is 0.999
\item \textbf{Rule 6} : If pH is $A_4$ and vfa is $A_3$ and Qin is
$A_3$ and $CH_4$ is $A_5$, then output is 1
\end{itemize}

Both rules indicate a high risk of acidogenesis with a high
$pH$, which is inconsistent with expert
knowledge of the acidogenic state.   Further investigation shows that each of these two rules is activated by only one
sample, which does not activate any other rule.
Indeed, one sample has a pH value of 8.5 (clearly not acid) and the other one has a pH of 7.6,
together with an alkalinity ratio (which should be low in an acidogenic state) greater than 0.8.

These two samples being labeled as erroneous data (maybe a sensor disfunction), we remove
them from the data set in further analysis.

This kind of outliers cannot be managed using traditional noise removal filtering techniques, it requires expert examination to decide whether they should be removed from learning data.

We renew the OLS procedure on the purified data, and we also perform a reduction of the output vocabulary, to improve interpretability.

\textbf{Performance with reduced output vocabulary \label{vocred}}

\begin{figure}[htbp]
\begin{tabular}[htbp]{ll}
\includegraphics[scale=0.4]{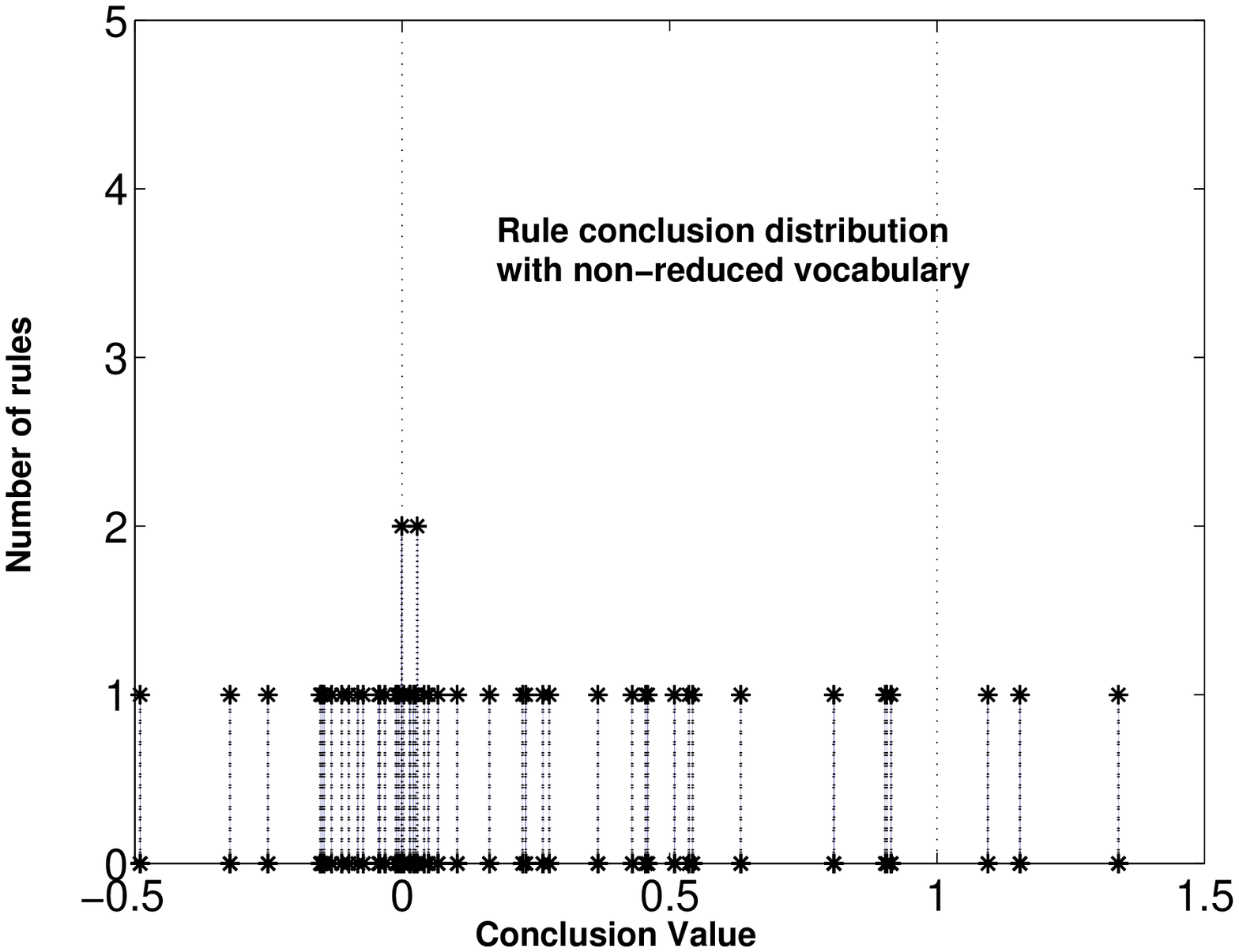} & \includegraphics[scale=0.4]{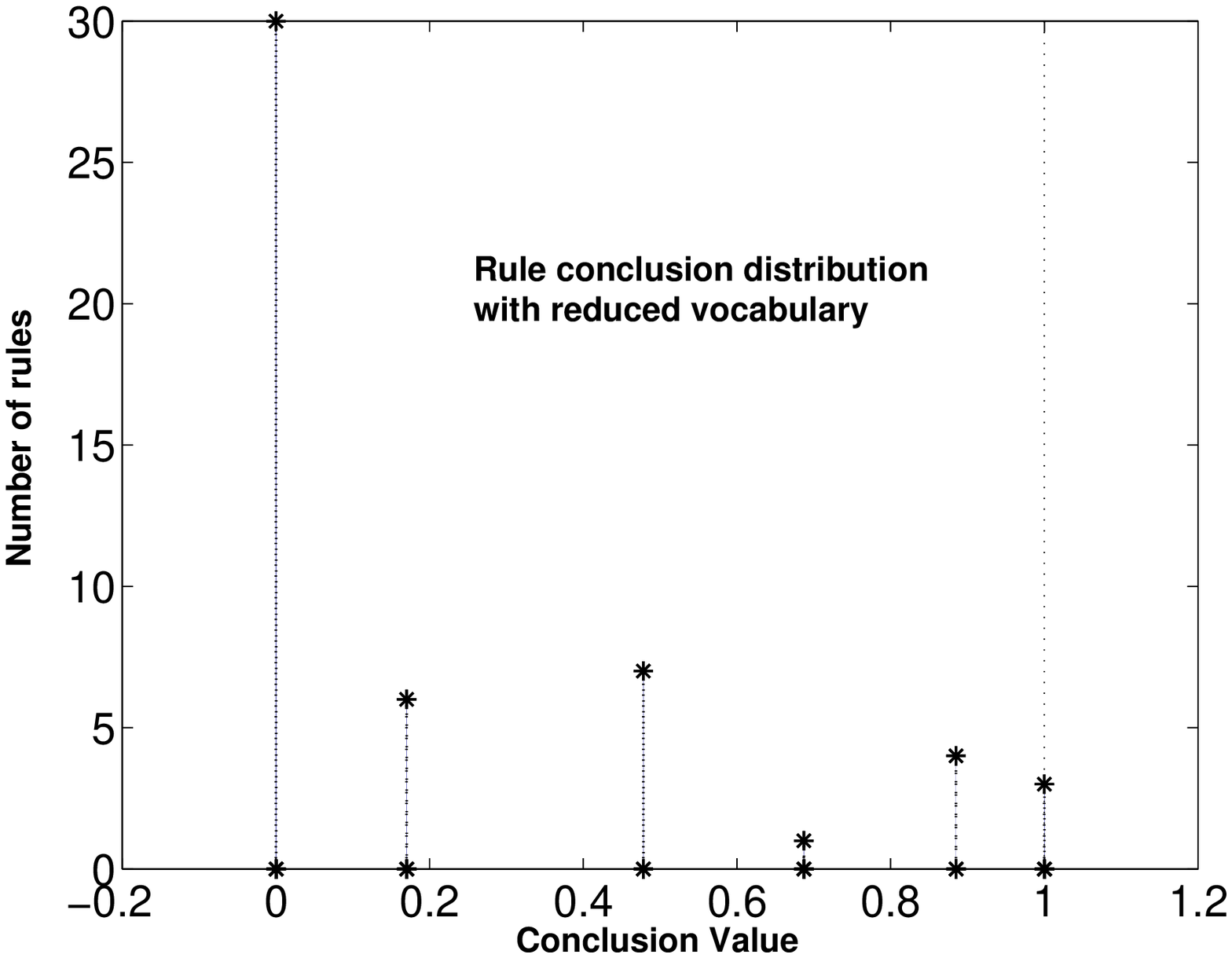}\\
\end{tabular}
\caption{Impact of vocabulary reduction on rule conclusions}
\label{fig:distribconc}
\end{figure}

The final rule base has 51 rules, the two rules induced by erroneous data having disappeared.
 The output vocabulary is reduced from 49 distinct values to 6 different ones, all of them constrained to belong to the output range.
Figure \ref{fig:distribconc} shows the rule conclusion distribution before and after vocabulary reduction. On the left subfigure, two dotted lines have been added to show the observed output range $[0-1]$. Rules are easier to interpret, while  the  distribution features are well conserved.
The new system performance is PI=0.056, which corresponds to an accuracy loss of 15 percent.


To test the rule base representativity, we did some experiments on increasing the activation threshold $\alpha$.
Up to $\alpha=0.5$, only one sample amongst the 587 ones is not covered by the rule base, which is a good sign as to
the robustness of our results.

Another interesting feature is that $100 \%$ of the
samples having an output greater than 0.2 are covered by the
first twenty rules, allowing one to first focus on this smaller set of
rules to describe critical states.

\begin{figure}[htbp]
\begin{center}
\includegraphics[scale=0.7]{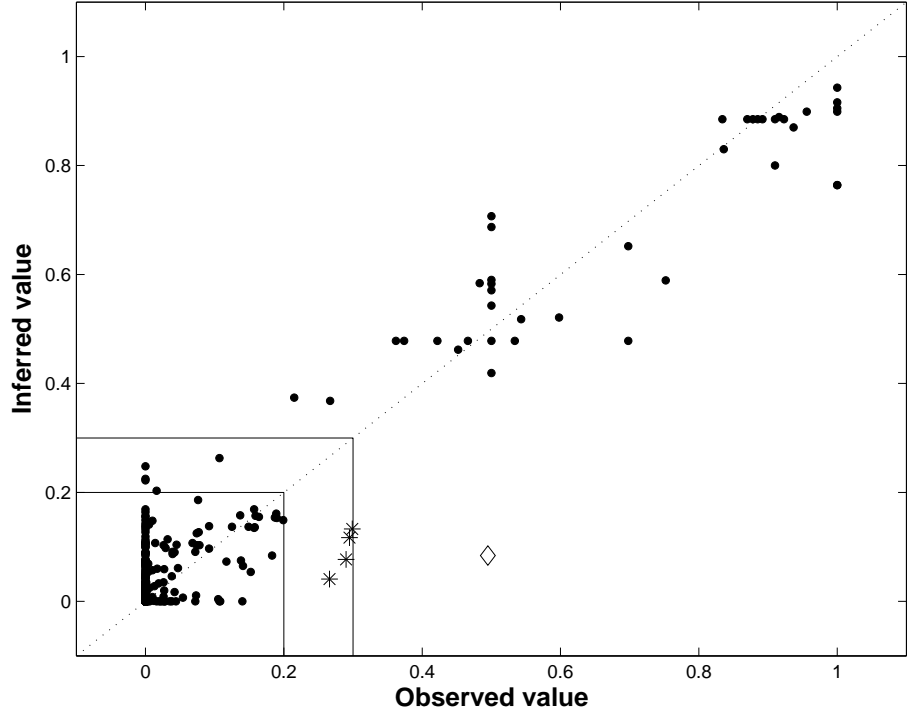}
\end{center}
\caption{Prediction with 6 conclusion values. $\bullet$ : Detection
with trigger $>$ 0.2 ; $\{ \ast, \diamondsuit \}$ : Non-detection with
trigger = 0.2 ; $\diamondsuit$ : Non-detection with trigger = 0.3}
\label{fig:Results}
\end{figure}

Figure \ref{fig:Results} illustrates the good qualitative predictive quality of the rule base: we can expect that the system
will detect a critical situation soon enough to prevent any collapse
of the process.
From a function approximation point of view, the prediction would be insufficient. However, for expert interpretation, figure \ref{fig:Results} is very interesting. Three  clusters appear. They can be labeled as \textit{Very low risk}, \textit{Non neglectable risk} and \textit{High risk}. They could be associated to three kinds of action or alarms.

From a fault detection point of view, some more time should be spent
on the few \textit{faulty} samples that wouldn't activate a fault detection
trigger set at 0.2 or 0.3. They have been signaled to experts for further investigation.
Each rule fired by those five samples (asterisk and diamond in
figure~\ref{fig:Results}) is also activated by about a hundred other
samples which have a very low acidogenic state. It may be difficult to draw conclusions from these five samples.

\section{Conclusion}
\label{section:conclusion}
Orthogonal transform methods are used to build compact rule base
systems. The OLS algorithm is of special interest as it takes into
account input data as well as output data.

Two modifications are proposed in this paper. The first one is related
to input partitioning. We propose to use a standardized fuzzy
partition with a reduced number of terms. This obviously improves
linguistic interpretability but also avoids the occurrence of an
important drawback of the OLS algorithm:  redundant rule selection.
Moreover, it can even enhance numerical accuracy.

The second way to improve linguistic interpretability is to deal with rule
conclusions. Reducing the number of distinct values used by the rules has some effect on the
numerical accuracy measured on the training sets, but very little impact on the performance obtained on test sets.

We have successfully applied the modified OLS to a
fault detection problem. Our results are robust, interpretable,
and our predictive capacity is more than acceptable. The OLS was also
shown able to detect some erroneous data after a first brief analysis.
When dealing with applications where the most important samples are
rare, OLS  can be very useful.

We would like to point out the double interest of properly used fuzzy concepts in a numerical technique. Firstly, linguistic reasoning with input data, which is only relevant with readable input partitions,  takes into account the progressiveness of biological phenomena which have a high intrinsic variability. Secondly, a similar symbolic reasoning can be used on output data. Though interesting for knowledge extraction, this is rarely considered.

Let us also underline the proposed modifications could benefit to all the
similar algorithms based on orthogonal transforms, for instance the TLS (Total Least Squares) method \cite{Yen99} which seems
to be of particular interest.

A thorough study of the robustness of this kind of models is still to be carried out. It should include a sensitivity analysis of both
algorithm parameters and data outliers with respect to the generalization ability.
The sensitivity analysis could be sampling-based or be based on statistical techniques (for instance decomposition of variance).
Similarly the rule selection procedure could be refined by extending classical backward-forward stepwise regression procedures to the fuzzy OLS algorithm.

Contrary to other methods, OLS does not perform a variable selection, which can be a serious drawback.
Future work should also focus on combining an efficient variable selection method with the OLS rule selection.

\end{document}